\documentclass[letterpaper, 10 pt, conference]{ieeeconf}  

\IEEEoverridecommandlockouts                              

\usepackage{cite}                               
\usepackage{amsmath,amssymb,amsfonts}           
\usepackage{graphicx}                           
\usepackage[table]{xcolor}                      
\usepackage{booktabs}                           
\usepackage{multirow}                           
\usepackage{makecell}                           
\usepackage{tabularx}                           
\usepackage{adjustbox}                          
\usepackage{pifont}                             
\usepackage{bbm} 
\usepackage{hyperref}

\definecolor{mygreen1}{HTML}{32CD32} 
\definecolor{mygreen2}{HTML}{90EE90} 
\definecolor{mygreen3}{HTML}{E2F0D9} 
\definecolor{myorange1}{HTML}{F4B183} 
\definecolor{myorange2}{HTML}{FBD4B4} 
\definecolor{myorange3}{HTML}{FFFACD} 
\definecolor{mypurple}{HTML}{E6E6FA}

\title{\LARGE \bf
QK-Bench: Benchmarking Quantitative Kinematic Understanding \\ for Dynamic Spatial Intelligence
}

\author{Lennox Wei$^{1}$, Jiaqi Yao$^{1}$, Shixin Sun$^{1}$, Yuanzheng Zhang$^{2}$, \\ Jianhua Han$^{3}$, Meng Tian$^{3,4}$, Shaohui Liu$^{1*}$, Wangmeng Zuo$^{1}$
\thanks{*Corresponding author: Shaohui Liu (e-mail: shliu@hit.edu.cn). }
\thanks{$^{1}$Lennox Wei, Jiaqi Yao, Shixin Sun, Shaohui Liu and Wangmeng Zuo are with School of Computer Science and Technology, Faculty of Computing, Harbin Institute of Technology, Harbin 150001, China.}%
\thanks{$^{2}$Yuanzheng Zhang is with University of Southern California, Los Angeles, CA 90089, USA.}
\thanks{$^{3}$Jianhua Han and Meng Tian are with Yinwang Intelligent Technology Co., Ltd., Shanghai 200135, China.}
\thanks{$^{4}$Meng Tian is with National University of Singapore, Singapore 119077, Singapore.}
}

\begin{document}

\maketitle
\thispagestyle{empty}
\pagestyle{empty}


\begin{abstract}
Spatial intelligence concerns both scene structure and its temporal evolution. Yet a semantically correct description can conceal a quantitative error: recognizing that objects are approaching does not determine their closing speed or whether they will meet. Such errors are critical in collision avoidance, trajectory tracking, and reactive grasping, where actions depend on motion magnitude, timing, and relative geometry. We study \textbf{quantitative kinematic understanding} in multimodal large language models (MLLMs): resolving motion magnitude, rate, reference-frame dependence, and future spatial evolution from visual evidence. We introduce the \textbf{Quantitative Kinematics Benchmark} (\textbf{QK-Bench}), comprising 2,341 multiple-choice QA pairs from 379 videos and \textbf{17 tasks} spanning \textbf{Object-Centric Kinematics}, \textbf{Camera Ego-Motion Kinematics}, and \textbf{Multi-Object Kinematic Relations}. Hidden geometric and temporal signals support trajectory construction and filtering, while fixed kinematic rules map physical quantities to verifiable answers, including categorical labels. Trajectory and question validation ensure target reliability and sufficient visual support in the sampled frames. All models receive a unified monocular RGB input. Evaluations of proprietary, open-source, and spatial MLLMs reveal substantial limitations: the strongest model achieves only 41.2\% overall accuracy, and spatial MLLMs show no consistent advantage. These findings indicate that current advances in spatial intelligence do not reliably transfer to quantitative kinematic understanding.
\end{abstract}

\section{Introduction}

Spatial intelligence enables multimodal large language models (MLLMs) to perceive and reason about the geometric and relational structure of the physical world~\cite{spatial457,3dsrbench,vsibench,spatialbench}. In dynamic environments, object motion, camera ego-motion, and multi-object interactions continually change that structure. A system may describe motion correctly while using incorrect quantities to decide when or where to act. In autonomous driving, recognizing an approaching vehicle does not establish its closing speed; underestimating this speed can make the time available for avoidance appear longer than it is. Manipulator trajectory tracking requires estimating target motion relative to the robot, whereas reactive grasping depends on the target position at the intended contact time. Errors in these estimates can cause tracking deviations or failed grasps of moving object. For robotic and embodied systems, the quantities omitted by a qualitative description can therefore determine whether an action is timely and spatially appropriate. These applications motivate evaluating the motion information on which downstream prediction, planning, and control depend.

\begin{figure}
    \vspace{+0.15cm}
    \centering
    \includegraphics[width=1.0\linewidth]{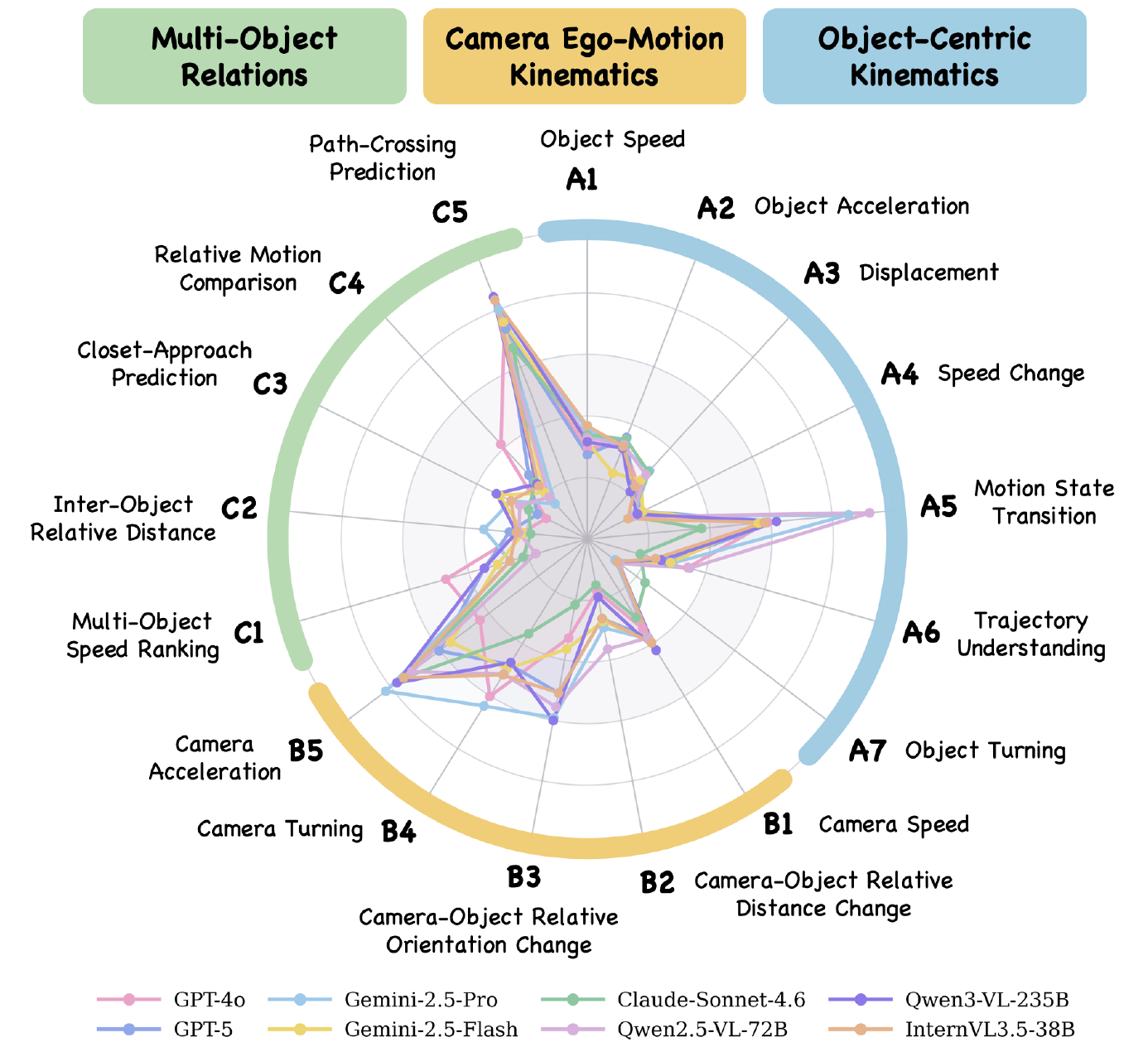}
    \vspace{-0.55cm}
    \caption{Task-wise performance of representative MLLMs on QK-Bench from monocular video inputs. The radar chart reports accuracy on 17 quantitative kinematic tasks spanning Object-Centric Kinematics, Camera Ego-Motion Kinematics, and Multi-Object Kinematic Relations, highlighting substantial capability differences across fine-grained task types.}
    \label{radar}
    \vspace{-0.55cm}
\end{figure}

\begin{figure*}
    \vspace{+0.15cm}
    \centering
    \includegraphics[width=1.0\linewidth]{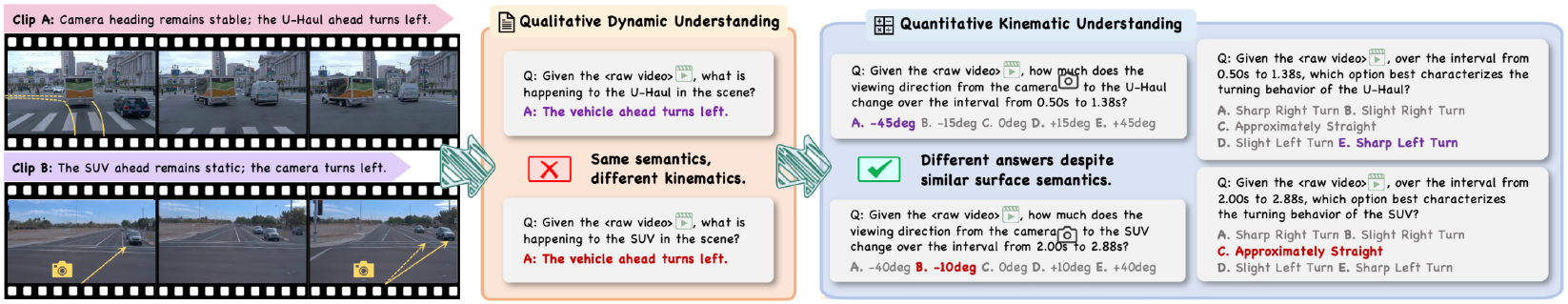}
    \vspace{-0.55cm}
    \caption{Similar image motion can conceal different underlying kinematics. The leftward image change arises from object motion in Clip A and primarily from camera ego-motion in Clip B. A correct qualitative description does not establish the object's displacement or speed in the relevant reference frame.}
    \label{semantic}
    \vspace{-0.55cm}
\end{figure*}

Recent benchmarks have extended spatial intelligence from static scenes to dynamic settings~\cite{stibench,vlm4d,dsibench,spatial4dbench} by assessing motion, viewpoint changes, and evolving spatial relations. This progress raises a further evaluation requirement: distinguishing motions that share a correct qualitative description but differ in quantitative properties and future outcomes. Two scenes may both show objects ``approaching'' yet differ in the time of closest approach; two trajectories may both be ``turning'' yet differ substantially in turning angle. A model can recognize either pattern correctly yet still misjudge these quantities. The distinction concerns the precision of spatial understanding, beyond the recognition of change alone.


We term this capability \textbf{quantitative kinematic understanding}. It extends spatial understanding from \textit{where} entities are and \textit{how} they change to \textit{how much} and \textit{how fast} they change, \textit{in which reference frame}, and \textit{what spatial configuration follows from current motion}. This spans individual object motion, camera ego-motion, camera--object relative dynamics, inter-object kinematic relations, and short-term prediction. Reliable evaluation requires constructing kinematic ground truth with consistent geometry and accurate timing. The resulting targets must distinguish motions that share a qualitative description but differ in magnitude, rate, or future outcome. Even categorical answers must be derived from measured quantities through fixed decision rules, so that selecting a plausible motion label alone is insufficient.


Figure~\ref{semantic} illustrates why reference frames matter for quantitative kinematic understanding. Similar leftward image changes arise from object motion in Clip A but primarily from camera ego-motion in Clip B. A model may correctly recognize this leftward change yet misjudge how far or how fast the object moves by attributing camera-induced changes to the object. Accurate quantitative judgments therefore require accounting for camera ego-motion and measuring change in the reference frame specified by the question.

We introduce the \textbf{Quantitative Kinematics Benchmark} (\textbf{QK-Bench}) to evaluate this capability. It comprises 2,341 question--answer pairs from 379 videos, organized into \textbf{17 tasks} spanning \textbf{Object-Centric Kinematics}, \textbf{Camera Ego-Motion Kinematics}, and \textbf{Multi-Object Kinematic Relations}. The two-stage pipeline uses hidden geometric and temporal signals to construct trajectories, compute physical quantities, and derive answers through fixed task-specific rules. Each target can therefore be traced to its source evidence and derivation. Trajectory filtering and question validation ensure both label reliability and the visual evidence needed to answer each question.

QK-Bench uses monocular RGB video as a common evaluation interface, consistent with existing spatial-intelligence benchmarks and applicable to heterogeneous MLLMs. Models receive RGB observations, corresponding timestamps, and natural-language questions. Geometric signals remain hidden during evaluation, enabling verifiable target construction while testing what models can infer from a shared visual input. As shown in Fig.~\ref{radar}, evaluations of proprietary, open-source, and spatial MLLMs reveal substantial limitations. The strongest model achieves 41.2\% overall accuracy, and the strongest open-source model achieves 39.1\%. Spatial MLLMs do not exhibit a consistent advantage in quantitative motion, camera--object relative dynamics, or future relational reasoning.
These findings indicate that current advances in spatial intelligence do not yet translate reliably into quantitative kinematic understanding.

Our contributions are as follows:
\begin{itemize}
    \item We formulate \textbf{quantitative kinematic understanding} as an evaluation objective for dynamic spatial intelligence: resolving motion magnitude, rate, reference-frame dependence, and future outcomes that qualitative descriptions leave unspecified.

    \item We introduce \textbf{QK-Bench}, with \textbf{17 tasks} spanning object motion, camera ego-motion, and multi-object relations. Its two-stage construction pipeline connects hidden geometric and temporal evidence to quantitative targets through trajectory validation and fixed kinematic rules.

    \item We evaluate proprietary, open-source, and spatial MLLMs, finding a best overall accuracy of 41.2\% and no consistent advantage for spatial MLLMs. The results identify quantitative kinematic understanding as a remaining limitation of current spatial intelligence.
\end{itemize}

\section{Related Work}

\subsection{Dynamic Spatial Intelligence Benchmarks}

Recent MLLM benchmarks have extended spatial intelligence from static scenes to temporally evolving environments. STI-Bench~\cite{stibench} evaluates object pose, displacement, motion estimation, and prediction, while DSI-Bench~\cite{dsibench} studies the coupled effects of camera self-motion and object motion. OmniSpatial~\cite{omnispatial}, VLM4D~\cite{vlm4d}, Dyn-Bench~\cite{thinkingindynamics}, and Spatial4D-Bench~\cite{spatial4dbench} further broaden dynamic spatial evaluation through viewpoint changes, object tracking, multi-object interactions, and temporal scene evolution. More recently, DSR-Bench~\cite{tencent} constructs dynamic spatial questions from geometric and motion cues, while 4DP-QA~\cite{4dpqa} explicitly addresses the entanglement of camera and object motion.

Prior work already includes motion estimation and prediction, with some benchmarks explicitly assessing numerical quantities. QK-Bench makes \textbf{quantitative kinematic understanding} the organizing objective of evaluation across object motion, camera ego-motion, and multi-object interactions. The required distinctions concern \emph{by how much} a configuration changes, \emph{how fast} it changes, \emph{relative to which reference frame}, and \emph{what spatial relation follows from the current motion}. These requirements concern the information needed to determine an answer, rather than its numerical or categorical format. For example, a turning category defined by fixed angular thresholds encodes a quantitative distinction even when the answer is expressed in words. QK-Bench uses such physically defined targets to assess whether models resolve differences in motion and future outcomes that qualitative descriptions can leave unspecified.

Table~\ref{comparison} separates dynamic scope from the quantities and predictions evaluated. The three task families organize this assessment; the central requirement is that answers resolve specified quantitative distinctions within each family.

\begin{table*}[htbp]
    \centering
    \vspace{+0.15cm}
    \caption{Comparison of dynamic scope and quantitative kinematic evaluation across representative benchmarks.}
    \label{comparison}
    \vspace{-0.25cm}
    \newcommand{\greencheck}{\textnormal{\textcolor{green!55!black}{\scalebox{1.18}{\ding{52}}}}}
    \newcommand{\redcross}{\textnormal{\textcolor{red!80!black}{\scalebox{1.26}{\ding{55}}}}}

    \setlength{\tabcolsep}{3.2pt}
    \renewcommand{\arraystretch}{1.12}

    \resizebox{0.95\textwidth}{!}{
    \begin{tabular}{l ccc cccc cc cc}
        \toprule[1.5pt]

        \multirow{3}{*}{\textbf{Benchmark}}
        & \multicolumn{3}{c}{\textbf{Dynamic Scope}}
        & \multicolumn{8}{c}{\textbf{Quantitative Kinematic Evaluation}}
        \\

        \cmidrule(lr{0.45em}){2-4}
        \cmidrule(lr{0.45em}){5-12}

        &
        \multirow{2}{*}{\textbf{Object}}
        & \multirow{2}{*}{\makecell[c]{\textbf{Camera}\\\textbf{Ego}}}
        & \multirow{2}{*}{\makecell[c]{\textbf{Multi-}\\\textbf{Object}}}

        & \multicolumn{4}{c}{\textbf{State Estimation}}
        & \multicolumn{2}{c}{\textbf{Relative Kinematics}}
        & \multicolumn{2}{c}{\textbf{Prediction}}
        \\

        \cmidrule(lr{0.35em}){5-8}
        \cmidrule(lr{0.35em}){9-10}
        \cmidrule(lr{0.35em}){11-12}

        &
        &
        &
        &
        \textbf{Disp.}
        & \textbf{Speed}
        & \textbf{Accel.}
        & \textbf{Turning}
        & \makecell[c]{\textbf{Cam.--}\\\textbf{Obj.}}
        & \makecell[c]{\textbf{Obj.--}\\\textbf{Obj.}}
        & \makecell[c]{\textbf{Closest}\\\textbf{Approach}}
        & \makecell[c]{\textbf{Path}\\\textbf{Crossing}}
        \\

        \midrule[1.0pt]

        STI-Bench~\cite{stibench}
        & \greencheck
        & \greencheck
        & \redcross
        & \greencheck
        & \greencheck
        & \greencheck
        & \greencheck
        & \redcross
        & \redcross
        & \redcross
        & \redcross
        \\

        DSI-Bench~\cite{dsibench}
        & \greencheck
        & \greencheck
        & \redcross
        & \redcross
        & \redcross
        & \redcross
        & \redcross
        & \redcross
        & \redcross
        & \redcross
        & \redcross
        \\

        QuantiPhy~\cite{quantiphy}
        & \greencheck
        & \redcross
        & \redcross
        & \redcross
        & \greencheck
        & \greencheck
        & \redcross
        & \redcross
        & \redcross
        & \redcross
        & \redcross
        \\

        DSR-Bench~\cite{tencent}
        & \greencheck
        & \redcross
        & \greencheck
        & \redcross
        & \redcross
        & \redcross
        & \redcross
        & \redcross
        & \redcross
        & \redcross
        & \redcross
        \\

        4DP-QA-Bench~\cite{4dpqa}
        & \greencheck
        & \greencheck
        & \greencheck
        & \redcross
        & \redcross
        & \redcross
        & \redcross
        & \redcross
        & \redcross
        & \redcross
        & \redcross
        \\

        \midrule[1.0pt]

        \textbf{QK-Bench (Ours)}
        & \textbf{\greencheck}
        & \textbf{\greencheck}
        & \textbf{\greencheck}
        & \textbf{\greencheck}
        & \textbf{\greencheck}
        & \textbf{\greencheck}
        & \textbf{\greencheck}
        & \textbf{\greencheck}
        & \textbf{\greencheck}
        & \textbf{\greencheck}
        & \textbf{\greencheck}
        \\

        \bottomrule[1.5pt]
    \end{tabular}
    }

    \vspace{-0.55cm}
\end{table*}

\subsection{Quantitative Kinematics and Motion Estimation}

Quantitative physical understanding has also received increasing attention in multimodal evaluation. QuantiPhy~\cite{quantiphy} evaluates numerical physical properties such as object size, velocity, and acceleration, showing that semantically plausible visual interpretations can remain quantitatively inaccurate. QK-Bench shares this concern with quantitative correctness and studies it within dynamic spatial intelligence. Its focus is how motion quantities determine the evolution of spatial configurations: individual object motion, reference-dependent camera--object dynamics, inter-object relations, and future spatial interactions. Numerical estimation and relational prediction are therefore evaluated as connected aspects of quantitative kinematic understanding.

Quantitative estimation of motion and geometry is fundamental to perception in robotics and autonomous systems. MonST3R~\cite{monst3r} and MegaSaM~\cite{megasam} recover dynamic scene geometry and camera motion from monocular videos, while SpatialTrackerV2~\cite{spatialtrackerv2} jointly models geometry, camera ego-motion, and object motion for world-space tracking. Recent 4D reconstruction methods such as Any4D~\cite{any4d} further recover metric geometry, camera poses, and dense 3D motion. Their explicit estimates provide geometric and temporal information for downstream localization, forecasting, planning, and interaction. QK-Bench examines whether MLLMs can infer quantitative kinematic information from visual observations and use it to answer questions about motion and its spatial consequences. Such estimators could serve as reference baselines on compatible task subsets, provided that their outputs and sensing assumptions are matched to the benchmark's target definitions.

\subsection{Models and Data for Dynamic Spatial Intelligence}

Progress in dynamic spatial intelligence has also been driven by specialized models, targeted supervision, and richer data resources. 4D-RGPT~\cite{4drgpt} introduces region-aware representations for dynamic scene understanding, while EgoDynamic4D~\cite{egodynamic4d} provides egocentric RGB-D observations, camera poses, and temporally consistent 4D annotations. MLLM-4D~\cite{mllm4d} improves the understanding of object motion, camera ego-motion, and camera--object relations through targeted data construction and post-training. Large-scale resources such as SpatialVID~\cite{spatialvid} and OmniWorld~\cite{omniworld} further provide diverse dynamic videos and geometric annotations for learning dynamic spatial representations.

These developments motivate a direct test of whether representations learned for spatial understanding support quantitatively correct judgments of motion, relative dynamics, and future interactions. Recognizing a changing spatial configuration does not determine the precision with which its evolution can be characterized. QK-Bench examines this transfer using targets with explicit physical definitions and a common visual input protocol for general-purpose and spatial MLLMs. The resulting assessment concerns information needed by embodied and autonomous systems, while keeping the evaluation focused on MLLM understanding.

\begin{figure*}
    \centering
    \vspace{+0.15cm}
    \includegraphics[width=0.99\linewidth]{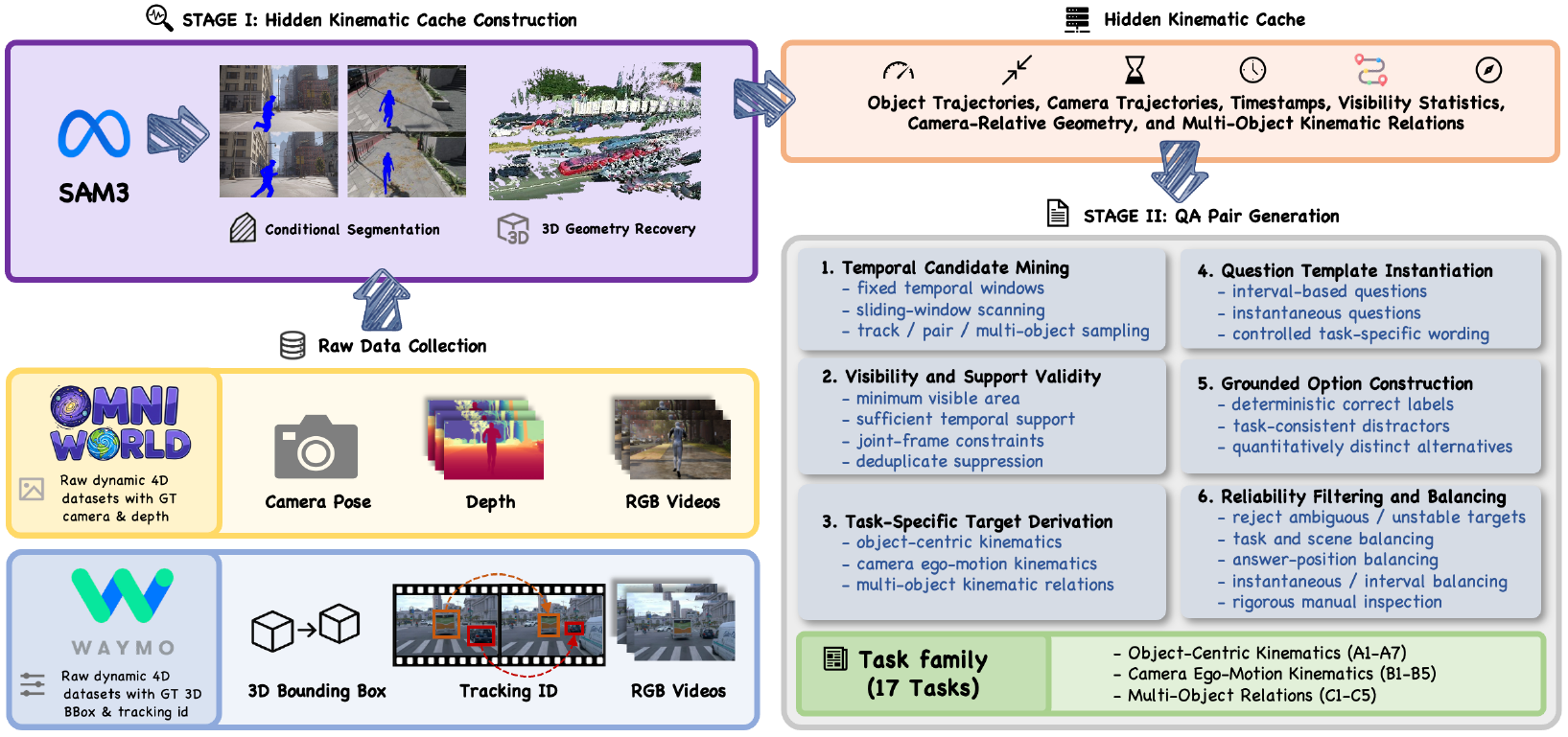}
    \vspace{-0.25cm}
    \caption{Overview of the QK-Bench construction pipeline. Stage I integrates geometric and temporal signals into a hidden kinematic cache, and Stage II derives deterministic quantitative targets and generates validated QA pairs through controlled option construction, reliability filtering, and balancing. Only monocular RGB frames, timestamps, and multiple-choice questions are exposed during evaluation.}
    \label{pipeline}
    \vspace{-0.45cm}
\end{figure*}

\section{QK-Bench}

\subsection{Benchmark Design and Task Taxonomy}

QK-Bench evaluates \textbf{quantitative kinematic understanding} through distinctions that qualitative motion descriptions leave unresolved. Its tasks require models to determine motion magnitude and rate, interpret changes in the specified reference frame, and infer future spatial relations from current motion. We make these requirements explicit in the targets: each answer is derived from a physically defined kinematic quantity using fixed task-specific rules. Quantitative grounding concerns the evidence needed to determine the answer, including when a continuous measurement is mapped to a categorical label.

The benchmark comprises 2,341 five-option multiple-choice QA pairs from 379 video clips drawn from Waymo~\cite{waymo} and OmniWorld~\cite{omniworld}. Its 17 tasks are systematically categorized into three distinct families. 
\textbf{Object-Centric Kinematics} covers Object Speed, Object Acceleration, Displacement, Speed Change, Motion State Transition, Trajectory Understanding, and Object Turning. \textbf{Camera Ego-Motion Kinematics} covers Camera Speed, Camera Acceleration, Camera Turning, Camera--Object Relative Distance Change, and Camera--Object Relative Orientation Change. \textbf{Multi-Object Kinematic Relations} covers Multi-Object Speed Ranking, Inter-Object Relative Distance, Relative Motion Comparison, Closest-Approach Prediction, and Path-Crossing Prediction under Current Motion Trend. Together, these families encompass quantitative state estimation, reference-dependent relative kinematics, and future interaction prediction. Interval-based questions assess motion over a specified window, whereas instantaneous questions assess local kinematic properties at a queried moment.

Construction separates \emph{target validity} from \emph{visual answerability}. Geometric and temporal signals support target derivation, while filtering and inspection assess whether the sampled RGB frames support the intended quantitative distinction. As illustrated in Fig.~\ref{pipeline}, Stage I integrates source signals into a temporally aligned hidden kinematic cache. Stage II derives task-specific targets from this cache and converts validated candidates into QA pairs. Privileged geometric signals are used only during construction and remain inaccessible during evaluation.

\subsection{Stage I: Hidden Kinematic Cache Construction}

\paragraph{Geometric and Temporal Signal Integration}
Quantifying motion requires positions to be compared in a consistent reference frame over known time intervals. Stage I therefore integrates heterogeneous source signals into a common temporally indexed representation. We parse RGB observations, native timestamps, depth maps, source annotations, and available camera metadata. Camera intrinsics and extrinsics define transformations among image, camera, and world coordinates. Motion quantities are computed using physical elapsed time from native timestamps, accounting for differences in source frame rates.

We use SAM 3~\cite{sam3} for prompt-conditioned segmentation of predefined object categories and back-project valid depth observations within each mask into 3D. Object positions are robustly estimated from the resulting points to reduce sensitivity to segmentation boundaries and local depth noise. Camera trajectories are obtained from pose metadata when available. The cache stores temporally aligned object positions, camera poses, timestamps, visibility flags, and representations in camera and world coordinates.

SAM 3 supplies localization cues only. Kinematic targets are computed after these observations have been combined with depth and camera metadata, temporally associated, and filtered. Motion magnitude and relative geometry are thus derived from the reconstructed trajectories and timestamps, rather than assigned by the segmentation model.

\paragraph{Temporal Association and Trajectory Filtering}
We associate frame-level observations using proximity in world and image coordinates, mask-area consistency, and temporal continuity. Association imposes limits on displacement and the number of missed frames. Short tracks and spurious motion caused by near-static jitter are filtered using temporal support and displacement. Duplicate detections are suppressed using trajectory proximity and image overlap.

\paragraph{Visibility and Temporal Support}
A target that can be computed from geometry may still lack sufficient evidence in the RGB input. We remove small object regions and check visibility using the same frame-sampling protocol as evaluation. For single-object questions, the queried entity must be visible in at least three sampled frames, including at least two consecutive frames. For multi-object questions, both trajectories must have valid observations in at least four temporally aligned frames. The two entities must also be jointly visible in at least three evaluation frames, including at least two consecutive frames.

\subsection{Stage II: Quantitative Target Derivation}

\paragraph{Candidate Mining and Kinematic Targets}
We scan the cache over task-specific intervals and retain candidates that meet the required trajectory, geometry, temporal support, and visibility criteria. Near-duplicate trajectory--query combinations are removed before question generation.

Kinematic quantities are computed in the reference frame specified by the question. Velocity and acceleration are obtained from timestamped trajectory differences, while displacement and relative distance are calculated from positions in the corresponding coordinates. Turning is measured from changes in planar motion direction. For categorical tasks, we first compute the continuous kinematic quantity and then apply the predefined task-specific decision rules. Labels such as ``slight turn'' and ``sharp turn'' therefore correspond to ranges of measured turning angles, making the distinction between them quantitative despite their verbal form.

Decision boundaries are fixed before QA generation and shared across all evaluated models. Speed Change, for example, uses thresholds of 0.20 and 0.60\,m/s to distinguish approximately constant, slight, and sharp changes in speed. Object Turning uses thresholds of $12^\circ$ and $25^\circ$, together with motion and trajectory-consistency checks, to exclude weak or unstable cases. Other categorical targets are derived from their respective quantities using fixed task-specific rules. For each retained instance, we record the source clip, queried interval, participating entities, reference frame, continuous target, and final label. These records allow each answer to be traced back to the trajectory and rule used to derive it. Label generation is independent of model predictions.

\subsection{Kinematic Prediction}

Approaching objects can differ in when they will be closest and whether they will occupy nearby positions at the same time. The predictive tasks assess these consequences of current motion using constant-velocity extrapolation. Their targets follow an explicit continuation assumption and do not account for later interactions or changes in motion.

Given the planar position $\mathbf{z}_i(t_e)\in\mathbb{R}^{2}$ of object $i$ at the end time $t_e$ of the observation window, its extrapolated position after an elapsed time $\Delta t \in[0,\infty)$ is
\begin{equation}
\tilde{\mathbf{z}}_i(\Delta t)=\mathbf{z}_i(t_e)+\mathbf{v}_i \cdot \Delta t ,
\end{equation}
where $\mathbf{v}_i\in\mathbb{R}^{2}$ denotes the planar velocity of object $i$ estimated from the hidden cache, and $\tilde{\mathbf{z}}_i(\Delta t)\in\mathbb{R}^{2}$ denotes its extrapolated planar position.

For \emph{Closest-Approach Prediction}, the extrapolated planar distance $d_{i,j}(\Delta t)\in\mathbb{R}_{0}^{+}$ between objects $i$ and $j$ is
\begin{equation}
d_{i,j}(\Delta t)=
\left\|
\tilde{\mathbf{z}}_i(\Delta t)
-
\tilde{\mathbf{z}}_j(\Delta t)
\right\|_2 ,
\end{equation}
and the time offset of closest approach is
\begin{equation}
\Delta t_{\mathrm{ca}}
=
\operatorname*{arg\,min}_{\Delta t}\:
d_{i,j}(\Delta t),
\end{equation}
where $\|\cdot\|_2$ denotes the Euclidean norm. The answer is determined from the closest-approach outcome or its corresponding task-specific interval.

For \emph{Path-Crossing Prediction under the Current Motion Trend}, the crossing event $y_{\mathrm{cross}}\in\{0,1\}$ occurs when the two extrapolated positions are within a specified tolerance at the same future time:
\begin{equation}
y_{\mathrm{cross}}
=
\mathbf{1}\!\left[
\exists \Delta t
\:\: \text{s.t.} \:\:
d_{i,j}(\Delta t) \le \epsilon
\right],
\end{equation}
where $\mathbf{1}[\cdot]$ is the indicator function and $\epsilon\in\mathbb{R}^{+}$ denotes the predefined planar crossing tolerance. Extrapolated outcomes are mapped to the task-specific answer categories. 
The crossing criterion requires the two extrapolated positions to be within $\epsilon$ at the same future time; passing through the same location at different times does not constitute a crossing.

\subsection{QA Construction and Label Reliability}

\paragraph{Physically Grounded Questions and Options}
We construct questions from validated targets using predefined task-specific templates for interval-based, instantaneous, relative, ranking, and predictive formulations. The templates explicitly specify the relevant entities and temporal query and control variation in wording. Distractors are derived from the same physical quantities or categories as the correct answer. Numerical options are generated by deterministic local discretization around the hidden target. Categorical options use fixed sets of physically defined states, ranking options use alternative object orderings, and predictive options use alternative outcomes or temporal categories. Options must be physically admissible for the task, with exactly one correct answer for the queried instance.

\paragraph{Reliability Control}
The trajectory and visibility checks precede target derivation. Candidates with unstable or ambiguous targets are excluded before question generation. We then inspect retained QA pairs for visual ambiguity and inconsistencies among the question, options, temporal evidence, and computed target. Fixed rules make label computation reproducible from a given trajectory; they do not by themselves establish the reliability of that trajectory or its visual support. Filtering and QA inspection address these separate sources of uncertainty before an instance enters final sampling. Only validated QA pairs enter final sampling. Correct-answer positions are approximately balanced across A--E to reduce positional bias.

\subsection{Evaluation Protocol}

For interval-based questions, we sample up to 24 RGB frames uniformly from the queried temporal window. For instantaneous questions, we use a local 9-frame window centered on the queried timestamp. All sampled frame timestamps are explicitly included in the prompt, and images are resized with the longest side capped at 512 pixels. The hidden cache, depth maps, camera poses, object trajectories, segmentation masks, and other privileged geometric signals are withheld from the models.

We adopt monocular RGB as a common visual interface, consistent with existing spatial-intelligence benchmarks. This shared visual input setting supports controlled comparisons of quantitative kinematic understanding across proprietary, open-source, and spatial MLLMs.


\begin{table*}[t]
\centering
\vspace{+0.15cm}
\caption{Zero-shot accuracy (\%) on QK-Bench.}
\vspace{-0.75cm}
\label{tab:combined_results}

\newcolumntype{Z}{>{\centering\arraybackslash}p{11pt}} 
\newcolumntype{M}{>{\centering\arraybackslash}p{14pt}} 
\newcolumntype{W}{>{\centering\arraybackslash}p{16pt}} 

\setlength{\tabcolsep}{1.2pt} 
\begin{center}
\resizebox{1.0\textwidth}{!}{
\fontsize{6}{7.5}\selectfont 
\begin{tabular}{l c c !{\vrule width 0.8pt} Z|Z|Z|Z|Z|Z|Z !{\vrule width 0.8pt} Z|Z|Z|Z|Z !{\vrule width 0.8pt} Z|Z|Z|Z|Z !{\vrule width 0.8pt} M|M|M !{\vrule width 0.8pt} W|W}
\toprule[1.5pt]
 & & & \multicolumn{7}{c !{\vrule width 0.8pt}}{\textbf{Object-Centric}} & \multicolumn{5}{c !{\vrule width 0.8pt}}{\textbf{Camera Ego-Motion}} & \multicolumn{5}{c !{\vrule width 0.8pt}}{\textbf{Multi-Object}} & \multicolumn{3}{c !{\vrule width 0.8pt}}{\textbf{Task Family}} & \multicolumn{2}{c}{\textbf{Capability}} \\
\cmidrule(l{2pt}r{2pt}){4-10} \cmidrule(l{2pt}r{2pt}){11-15} \cmidrule(l{2pt}r{2pt}){16-20} \cmidrule(l{2pt}r{2pt}){21-23} \cmidrule(l{2pt}r{2pt}){24-25}
\textbf{Methods} & \textbf{R} & \textbf{Ovr.} & \textbf{A1} & \textbf{A2} & \textbf{A3} & \textbf{A4} & \textbf{A5} & \textbf{A6} & \textbf{A7} & \textbf{B1} & \textbf{B2} & \textbf{B3} & \textbf{B4} & \textbf{B5} & \textbf{C1} & \textbf{C2} & \textbf{C3} & \textbf{C4} & \textbf{C5} & \textbf{Obj.} & \textbf{Cam.} & \textbf{Mult.} & \textbf{Und.} & \textbf{Pred.} \\
\midrule[1pt]
\rowcolor{mypurple} \multicolumn{25}{c}{\textbf{\textit{Proprietary Models (API)}}} \\
GPT-4o~\cite{gpt4o} & 3 & 34.7 & 29.5 & 32.9 & 24.3 & 18.2 & 58.8 & 34.0 & 12.3 & 34.2 & 17.2 & 32.7 & 60.2 & 43.7 & \cellcolor{myorange2}47.8 & 21.5 & 14.9 & \cellcolor{myorange2}41.7 & 74.4 & 29.8 & 42.4 & 38.0 & 34.8 & 46.8 \\
GPT-5~\cite{gpt5} & \cellcolor{mygreen3}2 & 35.2 & 27.4 & \cellcolor{myorange2}35.4 & 22.0 & 20.1 & 61.8 & 26.9 & 13.2 & 37.7 & 26.3 & 50.9 & 48.0 & 60.3 & 34.8 & 25.4 & 18.1 & 28.2 & 73.3 & 28.9 & 46.1 & 35.7 & 35.0 & 48.0 \\
Gemini-2.5-Pro~\cite{gemini2.5} & \cellcolor{mygreen1}1 & \cellcolor{myorange2}41.2 & 35.5 & 31.7 & \cellcolor{myorange2}30.1 & 16.7 & \cellcolor{myorange3}85.3 & 28.8 & 11.3 & 38.4 & 29.3 & \cellcolor{myorange3}59.1 & \cellcolor{myorange2}63.8 & \cellcolor{myorange2}82.1 & 26.1 & \cellcolor{myorange2}33.8 & 27.7 & 15.5 & 80.0 & \cellcolor{myorange3}33.6 & \cellcolor{myorange2}57.4 & 37.8 & \cellcolor{myorange2}41.4 & 55.5 \\
Gemini-2.5-Flash~\cite{gemini2.5} & 4 & 34.1 & 31.6 & 23.0 & 26.0 & 20.1 & 55.9 & 28.2 & 13.2 & 38.4 & 27.3 & 36.4 & 49.6 & 55.6 & 30.4 & 20.8 & 31.9 & 20.4 & 75.6 & 28.0 & 43.8 & 35.4 & 33.0 & 56.1 \\
Claude Sonnet 4.6~\cite{claude46} & 5 & 31.4 & 33.8 & \cellcolor{myorange3}34.8 & \cellcolor{myorange3}29.5 & 16.3 & 37.3 & 17.9 & \cellcolor{myorange2}23.6 & 30.1 & 15.2 & 21.8 & 36.2 & 74.8 & 21.7 & 18.5 & 21.3 & 25.2 & 66.7 & 28.8 & 37.9 & 31.2 & 31.2 & 45.7 \\
\midrule
\rowcolor{mypurple} \multicolumn{25}{c}{\textbf{\textit{Open-source Models}}} \\
Qwen2.5-VL-7B~\cite{qwen25} & 6 & 36.7 & 35.9 & 31.1 & 23.7 & 15.3 & 44.1 & 34.0 & 12.3 & 33.6 & 32.3 & 42.7 & 60.6 & 62.9 & 26.1 & 23.8 & 20.2 & 33.0 & 83.3 & 29.4 & 49.6 & 38.2 & 36.7 & 53.8 \\
Qwen2.5-VL-32B~\cite{qwen25} & 9 & 32.8 & 28.6 & 28.6 & 23.1 & 14.4 & 44.1 & \cellcolor{myorange3}35.9 & 16.0 & 39.0 & 10.1 & 30.9 & 57.5 & 45.0 & 21.7 & 20.0 & 24.5 & 22.3 & 83.3 & 28.1 & 41.5 & 35.2 & 32.2 & 56.1 \\
Qwen2.5-VL-72B~\cite{qwen25} & \cellcolor{mygreen1}1 & \cellcolor{myorange3}39.1 & 32.9 & 32.9 & 28.3 & 16.3 & \cellcolor{myorange2}92.2 & 34.6 & 13.2 & 37.7 & \cellcolor{myorange2}36.4 & 55.5 & 51.6 & 71.5 & 17.4 & 22.3 & 24.5 & 18.4 & 83.3 & \cellcolor{myorange2}34.9 & \cellcolor{myorange3}51.5 & 34.7 & \cellcolor{myorange3}39.2 & 56.1 \\
Qwen3-VL-8B~\cite{qwen3} & 4 & 37.0 & 31.6 & 33.5 & 19.7 & \cellcolor{myorange2}23.4 & 50.0 & 28.8 & 8.5 & 33.6 & 23.2 & 39.1 & 53.9 & 72.2 & 17.4 & 24.6 & \cellcolor{myorange2}36.2 & 38.8 & \cellcolor{myorange3}86.7 & 27.7 & 47.5 & \cellcolor{myorange3}42.9 & 35.3 & \cellcolor{myorange2}62.4 \\
Qwen3-VL-32B~\cite{qwen3} & 7 & 36.4 & \cellcolor{myorange2}38.0 & 33.5 & 23.1 & 20.1 & 49.0 & 28.2 & 13.2 & 36.3 & 13.1 & 43.6 & 49.6 & \cellcolor{myorange3}80.8 & 26.1 & 20.8 & 21.3 & 28.2 & 82.2 & 30.1 & 47.6 & 36.4 & 35.9 & 54.3 \\
Qwen3-VL-235B~\cite{qwen3} & 3 & 37.1 & 31.6 & 31.7 & 20.8 & 18.2 & 61.8 & 25.0 & 12.3 & \cellcolor{myorange2}42.5 & 19.2 & \cellcolor{myorange2}60.0 & 47.2 & 77.5 & 34.8 & 23.1 & 33.0 & 24.3 & 84.4 & 28.4 & 50.5 & 39.2 & 36.1 & \cellcolor{myorange3}60.7 \\
InternVL2.5-8B~\cite{internvl25} & 11 & 29.6 & 31.2 & 22.4 & 19.7 & 19.6 & 28.4 & 28.8 & 9.4 & 30.1 & 16.2 & 8.2 & 57.5 & 37.1 & 39.1 & 22.3 & 25.5 & 26.2 & 73.3 & 23.9 & 35.7 & 35.9 & 28.5 & 51.5 \\
InternVL2.5-38B~\cite{internvl25} & 8 & 35.3 & 30.8 & 25.5 & 24.3 & 16.7 & 70.6 & 32.7 & 13.2 & 35.6 & 28.3 & 52.7 & 55.9 & 45.7 & 26.1 & 21.5 & 23.4 & 21.4 & 82.2 & 29.8 & 45.9 & 35.2 & 34.8 & 54.9 \\
InternVL3.5-8B~\cite{internvl35} & \cellcolor{mygreen3}2 & 37.3 & 27.8 & 28.0 & 22.5 & 15.8 & 80.4 & 29.5 & 8.5 & 39.0 & 29.3 & 34.5 & 61.0 & 57.6 & 30.4 & 28.5 & 29.8 & 37.9 & 85.6 & 29.3 & 48.2 & \cellcolor{myorange2}43.1 & 36.7 & 59.0 \\
InternVL3.5-38B~\cite{internvl35} & 5 & 36.9 & \cellcolor{myorange3}36.8 & 32.3 & 23.1 & 14.8 & 57.8 & 23.1 & 12.3 & \cellcolor{myorange3}39.7 & 26.3 & 50.9 & 52.0 & 74.8 & 26.1 & 23.1 & 27.7 & 23.3 & 83.3 & 29.5 & 50.7 & 37.3 & 36.6 & 57.8 \\
LLaVA-OV-1.5-4B~\cite{llavaov} & 12 & 29.6 & 32.5 & 28.0 & 20.2 & 22.0 & 15.7 & 25.0 & 6.6 & 28.8 & 17.2 & 9.1 & 62.2 & 32.5 & \cellcolor{myorange3}43.5 & 18.5 & 33.0 & \cellcolor{myorange3}39.8 & 52.2 & 22.6 & 36.3 & 35.4 & 28.5 & 44.5 \\
LLaVA-OV-1.5-8B~\cite{llavaov} & 10 & 32.4 & 35.0 & 31.1 & 22.0 & \cellcolor{myorange3}23.0 & 15.7 & 26.3 & 9.4 & 30.1 & 18.2 & 16.4 & 61.8 & 41.1 & 21.7 & 20.8 & \cellcolor{myorange3}35.1 & 35.0 & 82.2 & 24.3 & 39.3 & 40.3 & 30.2 & \cellcolor{myorange3}60.7 \\
\midrule
\rowcolor{mypurple} \multicolumn{25}{c}{\textbf{\textit{Spatial MLLMs}}} \\
SpaceR-7B~\cite{spacer} & \cellcolor{mygreen3}2 & 35.5 & 32.1 & 26.7 & 20.2 & 16.7 & 67.6 & 34.6 & 8.5 & 33.6 & 27.3 & 37.3 & 60.2 & 64.2 & 21.7 & 23.8 & 17.0 & 17.5 & 82.2 & 29.3 & 48.3 & 33.6 & 35.3 & 52.0 \\
VST-7B-RL~\cite{vst} & 4 & 28.0 & 26.5 & 24.2 & 20.2 & 15.8 & 20.6 & 30.8 & 8.5 & 27.4 & 21.2 & 31.8 & 51.2 & 43.0 & 26.1 & 16.9 & 25.5 & 18.4 & 51.1 & 22.5 & 38.3 & 26.6 & 27.8 & 38.7 \\
Spatial-SSRL-7B~\cite{spatialssrl} & \cellcolor{mygreen1}1 & 36.7 & 35.9 & 31.1 & 19.1 & 22.0 & 39.2 & 34.6 & 7.5 & 32.9 & \cellcolor{myorange3}35.4 & 43.6 & 59.4 & 65.6 & 21.7 & 26.9 & 18.1 & 28.2 & 84.4 & 27.9 & 50.1 & 37.3 & 36.0 & 52.6 \\
SpatialLadder-3B~\cite{spatialladder} & 5 & 25.9 & 21.4 & 32.3 & 19.7 & 13.9 & 7.8 & 15.4 & 12.3 & 25.3 & 26.3 & 48.2 & 47.2 & 41.7 & 30.4 & 26.2 & 28.7 & 22.3 & 7.8 & 19.3 & 39.3 & 22.1 & 27.4 & 17.9 \\
Spatial-MLLM~\cite{spatialmllm} & 7 & 21.9 & 19.7 & 24.8 & 23.1 & 13.4 & 15.7 & 21.8 & \cellcolor{myorange3}20.8 & 16.4 & 22.2 & 26.4 & 28.7 & 33.1 & 17.4 & 19.2 & 17.0 & 36.9 & 6.7 & 20.6 & 26.1 & 19.8 & 23.3 & 10.4 \\
MLLM-4D-RFT~\cite{mllm4d} & 6 & 25.2 & 22.6 & 19.3 & 11.6 & 15.3 & 53.9 & \cellcolor{myorange2}37.8 & 16.0 & 26.7 & 29.3 & 4.5 & 30.3 & 12.6 & 17.4 & 28.5 & 18.1 & 15.5 & \cellcolor{myorange2}87.8 & 23.9 & 22.2 & 34.5 & 23.1 & 52.6 \\
VG-LLM~\cite{vgllm} & 8 & 21.6 & 22.6 & 19.3 & 12.7 & 22.5 & 45.1 & 22.4 & 13.2 & 24.7 & 26.3 & 9.1 & 37.4 & 5.3 & 30.4 & \cellcolor{myorange3}29.2 & 18.1 & 17.5 & 2.2 & 21.1 & 23.0 & 18.9 & 22.3 & 10.4 \\
DSR\_Suite-Model~\cite{tencent} & 3 & 32.4 & 35.5 & 27.3 & 22.5 & 21.1 & 14.7 & 27.6 & 10.4 & 28.8 & 30.3 & 28.2 & \cellcolor{myorange3}63.4 & 43.0 & 17.4 & 24.6 & 28.7 & 37.9 & 54.4 & 24.5 & 43.3 & 35.2 & 32.1 & 43.9 \\
\bottomrule[1.5pt]
\end{tabular}
}
\par\vspace{6pt} 
\parbox{\textwidth}{\scriptsize
\textit{Notes:}
Tasks are grouped into \textit{Object-Centric Kinematics} (A1--A7),
\textit{Camera Ego-Motion Kinematics} (B1--B5), and
\textit{Multi-Object Kinematic Relations} (C1--C5).
``Ovr.'' denotes overall accuracy; ``Obj.'', ``Cam.'', and ``Mult.''
denote task-family averages.
Orange cells mark the \colorbox{myorange2}{best} and
\colorbox{myorange3}{second-best} results across all models in each column.
``R'' ranks models within each category by overall accuracy, with
\colorbox{mygreen1}{first} and \colorbox{mygreen3}{second} places highlighted.
``Und.'' and ``Pred.'' denote understanding and predictive tasks, respectively.
All accuracies are reported as percentages with the percent sign omitted.
}

\vspace{-0.65cm}
\end{center}
\end{table*}

\section{Experiments}
\subsection{Evaluation Setup}
We conduct a zero-shot evaluation of proprietary, open-source, and spatially MLLMs, with results summarized in Table~\ref{tab:combined_results}. Each model receives a monocular RGB clip, a question, and five answer options; privileged geometric signals are withheld. Because each answer is derived from fixed kinematic rules, accuracy measures agreement with a quantitative target. No task-specific fine-tuning is performed. We report overall, task-wise, and family-level accuracy to examine performance across quantitative distinctions.

\subsection{Main Results}

\paragraph{Overall Performance}
Gemini-2.5-Pro~\cite{gemini2.5} achieves the highest overall accuracy, 41.2\%, followed by the strongest open-source model, Qwen2.5-VL-72B~\cite{qwen25}, at 39.1\%. Spatial-SSRL-7B~\cite{spatialssrl} leads spatial MLLMs at 36.7\%. These scores exceed the 20\% uniform-random reference, but even the overall leader answers fewer than half the questions correctly. Spatial MLLMs show no consistent advantage, despite leading on individual tasks. Spatial specialization therefore does not reliably translate into quantitative kinematic understanding across motion magnitude, rate, reference-frame dependence, and future evolution.


\paragraph{Quantitative Distinctions}
Object Turning and Speed Change remain challenging across models, with best accuracies of only 23.6\% and 23.4\%, respectively. Even categorical answers require quantitative judgments: distinguishing ``Slight Turn'' from ``Sharp Turn'' requires estimating the turning angle and comparing it with fixed thresholds. A model may correctly recognize a turn yet misjudge its magnitude and select the wrong category. These tasks evaluate whether models resolve differences in motion magnitude that a correct qualitative description leaves unspecified.

The best accuracies on Object Speed, Object Acceleration, and Displacement are 38.0\%, 35.4\%, and 30.1\%, respectively, revealing limitations across multiple aspects of quantitative kinematic understanding. Accurate estimates of these quantities are essential for deciding when to brake, where to intercept a moving object, and how to adjust a manipulator trajectory. A model may correctly recognize that an object is approaching yet misestimate its closing speed or future position. Such errors can leave an action mistimed or misplaced despite a correct qualitative interpretation.

\begin{figure*}
    \vspace{+0.15cm}
    \centering
    \includegraphics[width=0.98\linewidth]{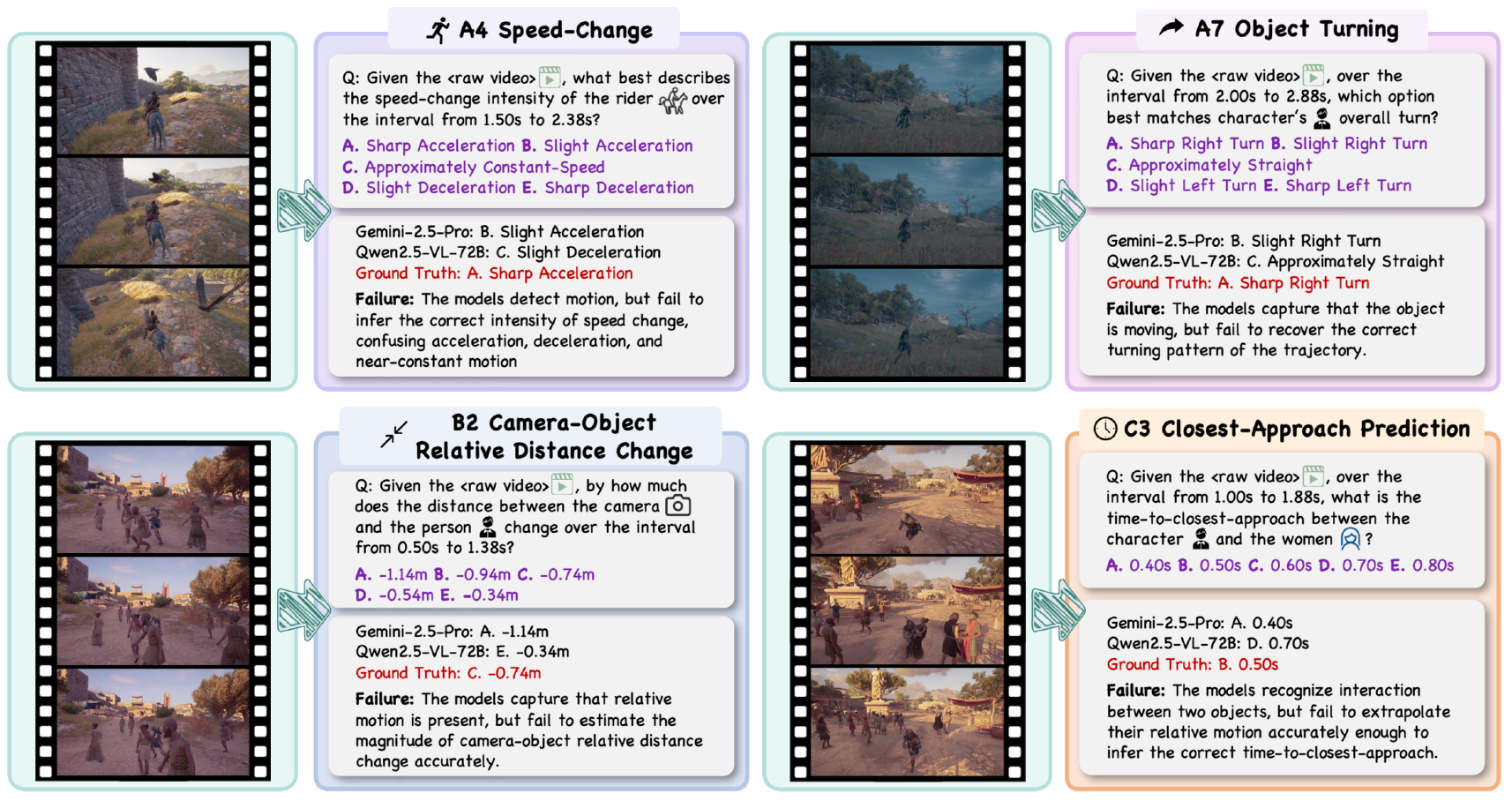}
    \vspace{-0.25cm}
    \caption{QK-Bench failure cases in Speed Change, Object Turning, Camera--Object Relative Distance Change, and Closest-Approach Prediction: plausible motion interpretations with incorrect quantitative judgments.}
    \label{failure}
    \vspace{-0.55cm}
\end{figure*}

\paragraph{Variation across Tasks}
No single model leads all three task families. Qwen2.5-VL-72B leads Object-Centric Kinematics with 34.9\%, Gemini-2.5-Pro leads Camera Ego-Motion Kinematics with 57.4\%, and InternVL3.5-8B leads Multi-Object Kinematic Relations with 43.1\%. Notably, even Gemini-2.5-Pro reaches only 33.6\% on Object-Centric Kinematics, illustrating how an overall score can obscure differences across families. Individual task leaders are distributed among nine models spanning all three categories. The overall ranking therefore combines different strengths in individual motion, relative dynamics, and prediction. Task-level diagnostics identify which quantitative requirements remain difficult for each model.

\paragraph{Predictive Understanding}
The highest prediction aggregate accuracies are 56.1\% for proprietary models and 62.4\% for open-source models. Performance nevertheless varies across individual tasks: MLLM-4D-RFT achieves 87.8\% on C5 but 52.6\% on the prediction aggregate. A strong result on one future-oriented task therefore does not establish consistent predictive understanding.

Recognizing that objects are approaching leaves their relative velocities and future timing unresolved. Closest-Approach Prediction asks when their separation is smallest, while Path-Crossing Prediction under Current Motion Trend requires evaluating whether their positions become sufficiently close at the same future time. Similar qualitative motion trends can thus lead to different answers. Both targets assume constant-velocity continuation and concern the specified consequences of current motion rather than behavior observed after the prediction window.


\paragraph{Failure Cases}
Fig.~\ref{failure} presents representative failure cases from Speed Change, Object Turning, Camera--Object Relative Distance Change, and Closest-Approach Prediction. These cases illustrate how a plausible qualitative interpretation can coexist with an incorrect quantitative judgment. Recognizing a turn or a speed change does not determine its magnitude; image displacement alone does not establish how camera--object distance changes; and apparent convergence does not determine the time of closest approach. By deriving answers from measured motion and explicit extrapolation rules, QK-Bench evaluates the quantitative differences that these qualitative descriptions leave unresolved.

\section{Discussion}

Spatial MLLMs show no consistent advantage, suggesting that spatial specialization alone does not reliably support quantitative kinematic understanding. A plausible explanation is that recognizing scene structure does not ensure sufficiently accurate scale, timing, and reference-frame alignment. These quantities matter for collision avoidance, manipulator trajectory tracking, and reactive grasping, even when motion is described correctly.

QK-Bench makes these requirements testable through targets derived from geometric and temporal evidence, validated trajectories, and fixed rules. This construction suggests a potential advantage for world models that reconstruct and update explicit 3D scene states. Estimated depth, camera poses, object masks, and temporal associations could form an inferred kinematic cache from RGB and timestamps. With sufficiently accurate scale and trajectories, the same rules could derive quantitative answers, including predictions under constant-velocity continuation. Compared with direct VLM-based answering, this paradigm could better support dynamic spatial intelligence by making motion quantities explicit and computable. QK-Bench provides a framework for testing whether this translates into more accurate and reliable quantitative kinematic understanding.

\section{Conclusion}

QK-Bench evaluates quantitative kinematic understanding within spatial intelligence through explicit targets for motion magnitude, rate, reference-frame dependence, and future evolution. Its construction links hidden geometric and temporal evidence to validated trajectories and fixed kinematic rules, making answers traceable. The best overall accuracy is 41.2\%, and spatial MLLMs show no consistent advantage. These findings motivate investigating world models with explicit scene reconstruction and kinematic computation. QK-Bench provides a testbed for assessing whether such representations can reliably recover the quantitative information needed to decide when and where to act.

\bibliographystyle{ieeetr}
\bibliography{method}

\end{document}